\documentclass[10pt,twocolumn,letterpaper]{article}

\usepackage{cvpr}
\usepackage{times}
\usepackage{epsfig}
\usepackage{graphicx}
\usepackage{amsmath}
\usepackage{amssymb}

\usepackage[utf8]{inputenc} 
\usepackage[T1]{fontenc}    
\usepackage{url}            
\usepackage{booktabs}       
\usepackage{amsfonts}       
\usepackage{nicefrac}       
\usepackage{microtype}      
\usepackage{amsthm}
\newtheorem{myTheo}{Lemma}
\usepackage{bbm}
\usepackage{graphicx}
\usepackage{subfigure}
\usepackage{multicol}  
\usepackage{multirow}

\usepackage[pagebackref=true,breaklinks=true,letterpaper=true,colorlinks,bookmarks=false]{hyperref}

\cvprfinalcopy 

\def\cvprPaperID{3524} 

\title{Rethinking Feature Distribution for Loss Functions in Image Classification}

\author{
Weitao Wan$^{1}$\thanks{These two authors contributed equally.}
\qquad Yuanyi Zhong$^{1,2*}$\thanks{This work was done when Y. Zhong was with Tsinghua University.}
\qquad Tianpeng Li$^{1}$
\qquad Jiansheng Chen$^{1}$\thanks{Corresponding author.} \\
$^1$Department of Electronic Engineering, Tsinghua University, Beijing, China \\
$^2$Department of Computer Science, University of at Urbana-Champaign, Illinois, USA\\
\texttt{wwt16@mails.tsinghua.edu.cn} \qquad \texttt{yuanyiz2@illinois.edu} \\
\texttt{ltp16@mails.tsinghua.edu.cn} \qquad \texttt{jschenthu@mail.tsinghua.edu.cn}
}

\begin{document}

\maketitle
\pagestyle{empty}  
\thispagestyle{empty} 

\begin{abstract}

We propose a large-margin Gaussian Mixture (L-GM) loss for deep neural networks in classification tasks.
Different from the softmax cross-entropy loss, our proposal is established on the assumption that the deep features of the training set follow a Gaussian Mixture distribution.
By involving a classification margin and a likelihood regularization, the L-GM loss facilitates both a high classification performance and an accurate modeling of the training feature distribution.
As such, the L-GM loss is superior to the softmax loss and its major variants in the sense that besides classification, it can be readily used to distinguish abnormal inputs, such as the adversarial examples, based on their features' likelihood to the training feature distribution.
Extensive experiments on various recognition benchmarks like MNIST, CIFAR, ImageNet and LFW, as well as on adversarial examples demonstrate the effectiveness of our proposal. 
\end{abstract}

\section{Introduction}

\begin{figure*}[t]
	\centering
	\includegraphics[width=0.75\linewidth]{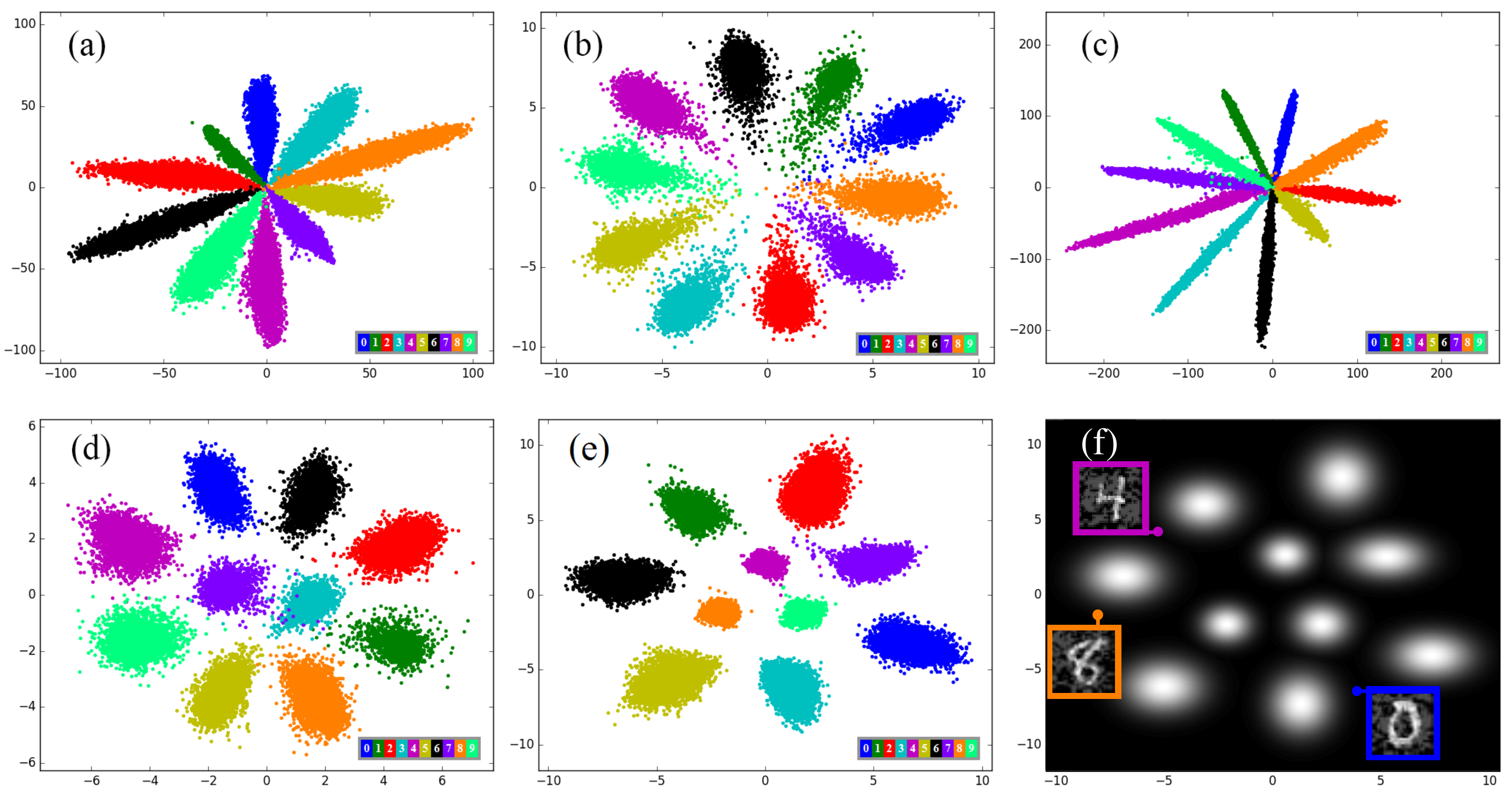} 
	\caption{Two-dimensional feature embeddings on MNIST training set. (a) Softmax loss. (b) Softmax loss + center loss \cite{{DBLP:conf/eccv/WenZL016}}. (c) Large-margin softmax loss \cite{DBLP:conf/icml/LiuWYY16}. (d) GM Loss without margin ($\alpha=0$). (e) Large-margin GM loss ($\alpha=1$). (f) Heatmap of the learned likelihood corresponding to (e). Higher values are brighter. Several adversarial examples generated by the Fast Gradient Sign Method \cite{goodfellow2014explaining} have extremely low likelihood according to the learned GM distribution and thus can be easily distinguished. This figure is best viewed in color.}
   \label{fig01}
\end{figure*}

Recently, deep neural networks have substantially improved the state-of-the-art performances of various challenging classification tasks, including image based object recognition \cite{DBLP:conf/nips/KrizhevskySH12, DBLP:conf/icml/IoffeS15, DBLP:conf/cvpr/HeZRS16}, face recognition \cite{DBLP:conf/cvpr/SchroffKP15, DBLP:conf/nips/SunCWT14} and speech recognition \cite{Dahl2011Context,Hinton2012Deep}.
In these tasks, the softmax cross-entropy loss, or the \emph{softmax loss} for short, has been widely adopted as the classification loss function for various deep neural networks \cite{DBLP:journals/corr/SzegedyVISW15, DBLP:conf/cvpr/HeZRS16, zagoruyko2016wide, larsson2016fractalnet, huang2017densely}.
For example in image classification, the affinity score of an input sample to each class is first computed by a linear transformation on the extracted deep feature. 
Then the posterior probability is modeled as the normalized affinity scores using the softmax function.
Finally, the cross-entropy between the posterior probability and the class label is used as the loss function.
The softmax loss has its probabilistic interpretation in that, for a large class of distributions, the posterior distribution complies with the softmax transformation of linear functions of the feature vectors \cite{christopher2006pattern}.
It can also be derived from a binary Markov Random Field or a Boltzmann Machine model \cite{deng2014large}.
However, the relationship between the affinity score and the probability distribution of the training feature space is vague.
In other words, for an extracted feature, its likelihood to the training feature distribution is not well formulated. 

Several variants have been proposed to enhance the effectiveness of the softmax loss. 
The Euclidean distances between each pair \cite{DBLP:conf/nips/SunCWT14} or among each triplet \cite{DBLP:conf/cvpr/SchroffKP15} of extracted features are added as an additional loss to the softmax loss.
Alternatively, in \cite{DBLP:conf/eccv/WenZL016} the Euclidean distance between each feature vector and its class centroid is used.
However, under the softmax loss formulation, the cosine distance based similarity metrics is more appropriate, indicating that using the Euclidean distance based additional losses may not be the most ideal choice.
Based on this understanding, an angular distance based margin is introduced in \cite{DBLP:conf/icml/LiuWYY16} to force extra intra-class compactness and inter-class separability, leading to better generalization of the trained models.
Nevertheless, the softmax loss is still indispensable and mostly dominates the training process in these proposals.
Therefore, the probabilistic modeling of the training feature space is still not explicitly considered.

In this paper we propose a Gaussian Mixture loss (GM loss) under the intuition that it is reasonable as well as tractable to assume the learned features of the training set to follow a Gaussian Mixture (GM) distribution, with each component representing a class.
As such, the posterior probability can be computed using the Bayes' rule.
The classification loss is then calculated as the cross-entropy between the posterior probability and the corresponding class labels.
To force the training samples to obey the assumed GM distribution, we further add a likelihood regularization term to the classification loss.
As such, for a well trained model, the probability distribution of the training features can now be explicitly formulated.
It can be observed from Fig.~\ref{fig01} that the learned training features spaces using the proposed GM loss are intrinsically different from those learned using the softmax loss and its invariants, by approximately following a GM distribution.

The GM loss is not just an alternative, it bears several essential merits comparing to the  softmax loss and its invariants. 
First, incorporating a classification margin into the GM loss is simple and straightforward so that there is no need to introduce an additional complicated distance function as is practiced in the large-margin softmax loss \cite{DBLP:conf/icml/LiuWYY16}.
Second, it can be proved that the center loss \cite{DBLP:conf/eccv/WenZL016} is formally equivalent to a special case of the likelihood regularization in the GM loss.
However, the classification loss and the regularization now share identical feature distance measurements in the GM loss since they are both induced from the same GM assumption.
Last but not the least, in addition to the classification result, the GM loss can be readily used to estimate the likelihood of an input to the learned training feature distribution, leading to the possibility of improving the model's robustness, for example, towards adversarial examples.

We discuss mathematic details of the GM loss in Section~\ref{method}.
Extensive experimental results on object classification, face verification and adversarial examples are shown in Section~\ref{exp}.
We conclude this work in Section~\ref{conclusions} .

\section{Related Work}

The previous efforts for overcoming certain deficiencies of the softmax loss are inspiring.
One of the most widely studied technical route is to explicitly encourage stronger intra-class compactness and larger inter-class separability while using the softmax loss.
Y. Sun \etal introduced the contrastive loss in training a Siamese network for face recognition by simultaneously minimizing the distances between positive face image pairs and enlarging the distances between negative face image pairs by a predefined margin \cite{DBLP:conf/nips/SunCWT14}.
Similarly, F. Schroff \etal proposed to apply such inter sample distance regularizations on image triplets rather than on image pairs \cite{DBLP:conf/cvpr/SchroffKP15}.
A major drawback of the contrastive loss and the triplet loss is the combinatoric explosion in the number of image pairs or triplets especially for large-scale data sets, leading to the significant increase in the required number of training iterations.
The center loss proposed in \cite{DBLP:conf/eccv/WenZL016} effectively circumvents the pair-wise or triplet-wise computation by minimizing the Euclidean distance between the features and the corresponding class centroids.
However, such a formulation brings about inconsistency of distance measurements in the feature space.
W. Liu \etal solved this problem by explicitly introduced an angular margin into the softmax loss through the designing of a sophisticated differentiable angular distance function \cite{DBLP:conf/icml/LiuWYY16}.
Another technical route mainly aims at improving the numerical stability of the softmax loss. 
Along this line, the label smoothing \cite{DBLP:journals/corr/SzegedyVISW15} and the knowledge distilling \cite{Hinton2015Distilling} are two typical methods of which the basic idea is to replace the one-hot ground truth distribution with other distributions that are probabilistically more reasonable.
An interesting recent work proposed by B. Chen \etal focused on mitigating the early saturation problem of the softmax loss by injecting annealed noise in the softmax function during each training iteration \cite{Chen2017Noisy}.
Generally speaking, all these works aim at improving the softmax loss rather than reformulating its fundamental assumption.

It has been revealed that deep neural networks with high classification accuracies are vulnerable to adversarial examples \cite{goodfellow2014explaining}.
Previous methods for solving this dilemma either directly included the adversarial samples in the training set \cite{kurakin2016adversarial} or introduced an new model for detecting the spoofing samples \cite{metzen2017detecting}.
Intuitively, however, the features of adversarial examples should follow a probability distribution quite different from that of the learned training feature space. 
In other words, it is possible to better distinguish the adversarial examples if the distribution of the training feature space can be explicitly modeled.

\section{Gaussian Mixture Loss}
\label{method}
In this section, we will formulate the GM loss from a probability perspective. 
We will also describe how to efficiently add a classification margin to the GM loss, after which the likelihood regularization term in the GM loss is further discussed.
The optimization of the GM loss is also presented.

\subsection{Intuitions}
\label{intuition}

Considering a $K$ class classification task in which the softmax loss is used.
For an input sample with $x$ as its extracted deep feature vector, its posterior probability of belonging to a certain class $j \in [1,K]$ can be expressed by Eq.~\ref{eq_softmax}, in which the affinity score (logit) $f_k(x)$ is usually calculated by linearly transforming the feature vector $x$ as is shown in Eq.~\ref{logit}.
In practice, the linear functions of all the $K$ classes are combined to form a linear transformation layer with all the $w_k, b_k$ as the trainable parameters.
A larger value of the affinity score $f_k(x)$ indicates a higher posterior probability of $x$ belonging to the class $k$.
However, $f_k(x)$ cannot be directly used to evaluate $x$'s likelihood to the distribution of the training features which is not explicitly formulated at all.
\begin{equation}
\label{eq_softmax}
{p}(j|x) = \frac{e^{f_j(x)}}{\sum_{k=1}^{K} e^{f_k(x)}}
\end{equation}
\begin{equation}
\label{logit}
f_k(x) = w_k^Tx + b_k, k \in [1,K]
\end{equation}

What is more, since $f_k(x)$ is computed through inner product, the similarity between features in the learned feature space should be measured using the cosine distance.
However, in the Euclidean distance based regularization is more widely adopted in softmax variants probably due to its mathematical simplicity.
For example, the Euclidean distance between the extracted feature and the corresponding class centroid was used to formulate the center loss $\mathcal{L}_C$ in Eq.~\ref{center_loss} \cite{DBLP:conf/eccv/WenZL016}, in which $N$ is the number of training samples; $x_i$ and $z_i$ are the extracted feature and the class label of the $i$-th sample respectively; and $\mu_{z_i}$ is the feature centroid (mean) for class $z_i$.
Intuitively, such a regularization should be more reasonable if the similarity measurement can be coherent to that in the classification loss.
\begin{equation}
\label{center_loss}
\mathcal{L}_C = \frac{1}{2}\sum_{i=1}^{N}\|x_i - \mu_{z_i}\|_2^2
\end{equation}

\subsection{GM loss formulation}
\label{sect_formulation}

Different from the softmax loss, we hereby assume that the extracted deep feature $x$ on the training set follows a Gaussian mixture distribution expressed in Eq.~\ref{rbfprob1}, in which $\mu_k$ and $\Sigma_k$ are the mean and covariance of class $k$ in the feature space; and $p(k)$ is the prior probability of class $k$.
\begin{equation}
\label{rbfprob1}
p(x) = \sum_{k=1}^{K} \mathcal{N}(x;\mu_k,\Sigma_k) p(k) 
\end{equation}

Under such an assumption, the conditional probability distribution of a feature $x_i$ given its class label $z_i \in [1,K]$ can be expressed in Eq.~\ref{rbfprob2}. Consequently, the corresponding posterior probability distribution can be expressed in Eq.~\ref{rbfprob3}.
\begin{equation}
\label{rbfprob2}
p(x_i|z_i) = \mathcal{N}(x_i;\mu_{z_i},\Sigma_{z_i})
\end{equation}
\begin{eqnarray}
\begin{aligned}
\label{rbfprob3}
p(z_i|x_i) = \frac{\mathcal{N}(x_i;\mu_{z_i},\Sigma_{z_i})p(z_i)}{\sum_{k=1}^{K}\mathcal{N}(x_i;\mu_k,\Sigma_k) p(k)}
\end{aligned}
\end{eqnarray}

As such, a \emph{classification loss} $\mathcal{L}_{cls}$ can be computed as the cross-entropy between the posterior probability distribution and the one-hot class label as is shown in Eq.~\ref{eq_lossi}, in which the indicator function $\mathbbm{1}()$ equals $1$ if $z_i$ equals $k$; or 0 otherwise.
\begin{eqnarray}
\label{eq_lossi}
\begin{aligned}
\mathcal{L}_{cls} &= -\frac{1}{N}\sum_{i=1}^{N}\sum_{k=1}^{K} \mathbbm{1}(z_i=k) \log {p}(k|x_i) \\
&= -\frac{1}{N}\sum_{i=1}^{N}\log \frac{\mathcal{N}(x_i;\mu_{z_i},\Sigma_{z_i})p(z_i)}{\sum_{k=1}^{K}\mathcal{N}(x_i;\mu_k,\Sigma_k) p(k)}
\end{aligned}
\end{eqnarray}

Optimizing the classification loss only cannot explicitly drive the extracted training features towards the GM distribution.
For example, a feature $x_i$ can be far away from the corresponding class centroid $\mu_{z_i}$ while still being correctly classified as long as it is relatively closer to $\mu_{z_i}$ than to the feature means of the other classes.
To solve this problem, we further introduce a \emph{likelihood regularization} term for measuring to what extent the training samples fit the assumed distribution.
The likelihood for the complete data set \{$X,Z$\} is expressed in Eq.~\ref{likelihood}.
We define the likelihood regularization term as the negative log likelihood shown in Eq.~\ref{likelihood2}.
By reasonably assuming constant prior probabilities $p(z_i)$, the likelihood regularization $\mathcal{L}_{lkd}$ can be simplified as Eq.~\ref{l_reg}.
\begin{equation}
\label{likelihood}
p(\mathrm{X,Z|\mu,\Sigma}) = \prod_{i=1}^N\prod_{k=1}^K\mathbbm{1}(z_i=k)\mathcal{N}(x_i;\mu_{z_i},\Sigma_{z_i})p(z_i)
\end{equation}
\begin{equation}
\label{likelihood2}
\log p(\mathrm{X,Z|\mu,\Sigma}) = -\sum_{i=1}^N(\log \mathcal{N}(x_i;\mu_{z_i},\Sigma_{z_i}) + \log p(z_i))
\end{equation}
\begin{equation}
\label{l_reg}
\mathcal{L}_{lkd} = -\sum_{i=1}^N\log \mathcal{N}(x_i;\mu_{z_i},\Sigma_{z_i})
\end{equation}

Finally the proposed GM loss $\mathcal{L}_{GM}$ is defined in Eq.~\ref{l_gm}, in which $\lambda$ is a non-negative weighting coefficient.
\begin{equation}
\label{l_gm}
\mathcal{L}_{GM} = \mathcal{L}_{cls} + \lambda \mathcal{L}_{lkd}
\end{equation}

By definition, for the training feature space, the classification loss $\mathcal{L}_{cls}$ is mainly related to its discriminative capability while the likelihood regularization $\mathcal{L}_{lkd}$ is related to its probabilistic distribution.
Under the GM distribution assumption, $\mathcal{L}_{cls}$ and $\mathcal{L}_{lkd}$ share all the parameters.

\subsection{Large-Margin GM Loss}

It has been widely recognized in statistical machine learning that large classification margin on the training set usually helps generalization, which is also believed to be applicable in deep learning \cite{DBLP:conf/aaai/SunCWLL16,DBLP:conf/icml/LiuWYY16}.
Denote $x_i$'s contribution to the classification loss to be $\mathcal{L}_{cls,i}$, of which an expansion form is in Eq.~\ref{l_cls_i} and Eq.~\ref{l_cls_d}.
\begin{equation}
\label{l_cls_i}
\mathcal{L}_{cls,i} = -\log \frac{p(z_i)|\Sigma_{z_i}|^{-\frac{1}{2}}e^{-d_{z_i}}}{\sum_{k}p(k)|\Sigma_k|^{-\frac{1}{2}}e^{-d_k}}
\end{equation}
\begin{equation}
\label{l_cls_d}
d_k = (x_i - \mu_k)^T\Sigma_{k}^{-1}(x_i - \mu_k) / 2
\end{equation}

Since the squared Mahalanobis distance $d_k$ is by definition non-negative, a classification margin $m \geq 0$ can be easily introduced to achieve the large-margin GM loss as in Eq.~\ref{l_cls_m}.
Obviously, adding the classification margin to the GM loss is more straightforward than to the softmax loss \cite{DBLP:conf/icml/LiuWYY16}.   
It should be emphasized that such a simple formulation cannot be directly applied to the softmax loss since an inner product can be negative, whereas a margin generally has to be non-negative to make sense.
\begin{equation}
\label{l_cls_m}
\mathcal{L}_{cls,i}^{m} = - \log \frac{p(z_i)|\Sigma_{z_i}|^{-\frac{1}{2}}e^{-d_{z_i}-m}}{\sum_{k}p(k)|\Sigma_k|^{-\frac{1}{2}}e^{-d_k-\mathbbm{1}(k=z_i)m}}
\end{equation}

To understand $m$'s role in the large-margin GM loss, one may consider the simplest case in which $p(k)$ and $\Sigma_k$ are identical for all the classes. 
Then $x_i$ is classified to the class $z_i$ if and only if Eq.~\ref{explain_margin} holds, indicating that $x_i$ should be closer to the feature mean of class $z_i$ than to that of the other classes by at least $m$.
\begin{equation}
\label{explain_margin}
e^{-d_{z_i} - m} > e^{-d_k} \Longleftrightarrow d_k - d_{z_i} > m \quad ,\forall k \ne z_i
\end{equation}

\begin{figure}[t]
	\centering
	\includegraphics[width=0.99\linewidth]{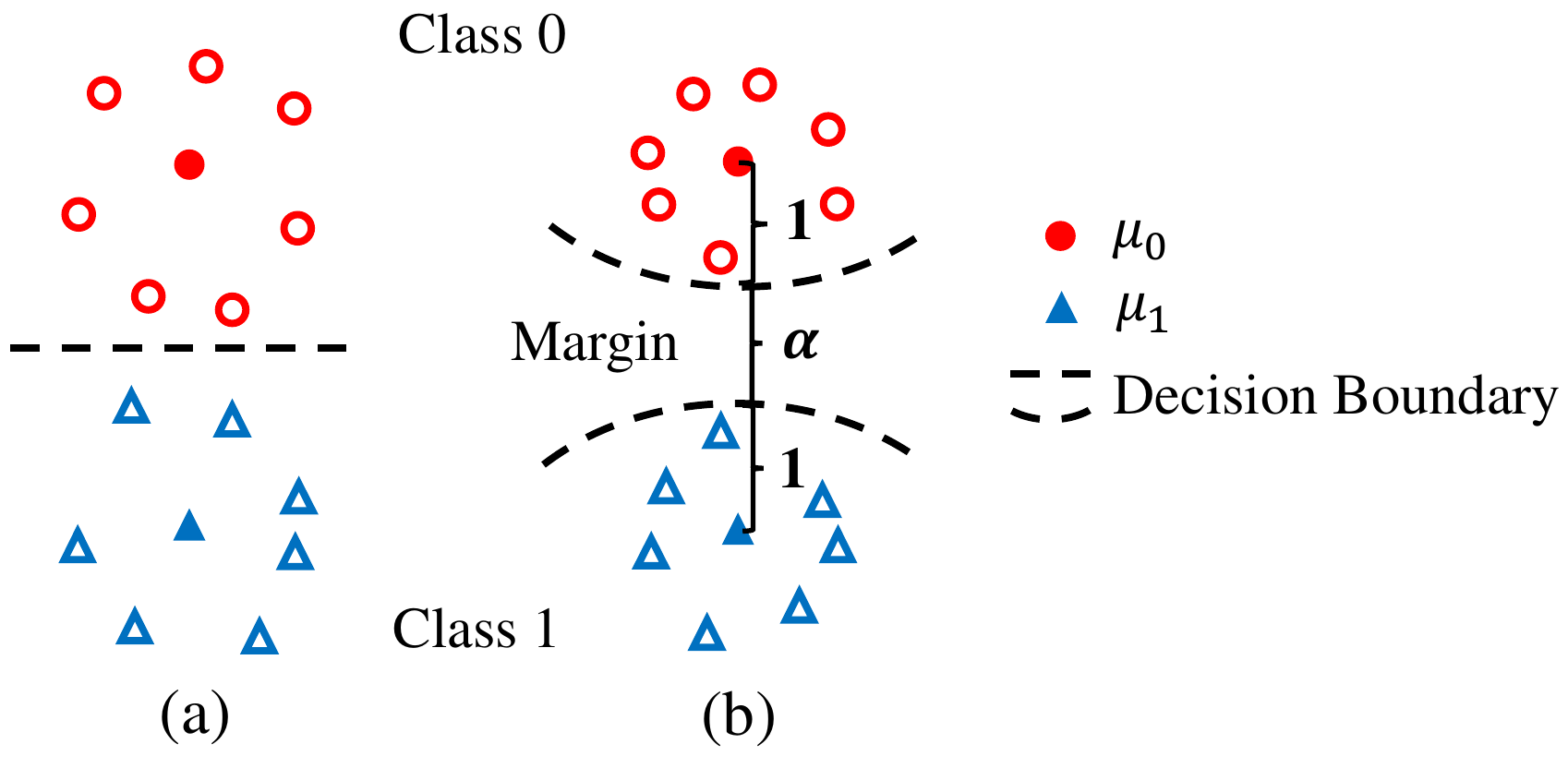}
	\caption{A geometry interpretation of the relationship between $\alpha$ and the margin size in the training feature space using (a) GM loss without margin $\alpha=0$; (b) large-margin GM loss with $\alpha>0$.} 
   \label{fig_geometry}
\end{figure}

To design the margin, we adopt an adaptive scheme by letting the value of $m$ to be proportional to each sample's distance to its corresponding class feature mean, i.e., $m = \alpha d_{z_i}$, in which $\alpha$ is a non-negative parameter controlling the size of the expected margin between two classes on the training set.
Fig.~\ref{fig_geometry} shows a schematic interpretation of $\alpha$; and Fig.~\ref{fig01} (d) and (e) illustrate how the training feature space changes when increasing $\alpha$ from $0$ to $1$.

\subsection{A Discussion on $\mathcal{L}_{lkd}$}
\label{discussion}

Although the likelihood regularization $\mathcal{L}_{lkd}$ defined in Eq.~\ref{l_reg} is proposed from a probability perspective, it has a strong connection with the empirical center loss $\mathcal{L}_C$ defined in Eq.~\ref{center_loss} \cite{DBLP:conf/eccv/WenZL016} as is described in \textbf{Lemma \ref{theo1}}, of which the proof is quite straightfoward.

\begin{myTheo}
\label{theo1}
If $\Sigma_k=I$ (identity matrix), $p(k)=1/K, \forall k \in [1,K]$, the center loss $\mathcal{L}_C$ and the likelihood regularization $\mathcal{L}_{lkd}$ satisfy Eq.~\ref{lc_rel}, in which $D$ is the feature dimension.
\begin{equation}
\label{lc_rel}
\mathcal{L}_{lkd} = \mathcal{L}_C + \frac{N}{2}D\log (2\pi)
\end{equation}
\end{myTheo}

\textbf{Lemma \ref{theo1}} shows that $\mathcal{L}_C$ is identical to $\mathcal{L}_{lkd}$ except for a constant under certain conditions.
In other words, the center loss \cite{DBLP:conf/eccv/WenZL016} is basically equivalent to a special case of the proposed likelihood regularization.
This indicates that it might be more appropriate to use the center loss, or the proposed likelihood $\mathcal{L}_{lkd}$ as regularization in a GM distributed feature space, as is practiced in this work.

More importantly, $\mathcal{L}_{lkd}$ can be readily used to estimate the likelihood of a sample feature to the learned GM distribution. 
Simply put, a model trained using our GM loss can now both generate a classification result and provide a likelihood estimation.
In case that the likelihood is too low, one may refuse to make the classification decision. 
Such a choice may be favorable, for example, when an adversarial example \cite{goodfellow2014explaining} is generated to attack the trained classification model.
In fact, the center loss $\mathcal{L}_C$ could also be used to estimate such a likelihood.
However, when being combined with the softmax loss during training, the center loss may produce inaccurate likelihood estimation since the generated training feature space probably deviates from the GM distribution. 



\subsection{Optimization}
\label{optimization}

The GM loss can be optimized using the typical stochastic gradient descent (SGD) algorithm.
In practice, updating the covariance matrix with gradient descent is feasible but may suffer from singularity problems.
Hence, for simplicity, we assume that the covariance matrix $\Sigma_k$ is diagonal, denoted by $\Lambda_k$; and the prior probability $p(k)=1/K$.
As such, the contribution of a sample $x_i$ to the large-margin GM loss can be rewritten in Eq.~\ref{l_gm_simple} and Eq.~\ref{distdiag}. 


\begin{eqnarray}
\begin{aligned}
\label{l_gm_simple}
\mathcal{L}_{GM,i}^m =& -\log \frac{|\Lambda_{z_i}|^{-\frac{1}{2}} e^{-d_{z_i}(1+\alpha)}}{\sum_{k} |\Lambda_{k}|^{-\frac{1}{2}}e^{-d_k(1+\mathbbm{1}(k=z_i)\alpha)}} \\
&+ \lambda (d_{z_i}+\frac{1}{2}\log |\Lambda_{z_i}|) 
\end{aligned}
\end{eqnarray}
\begin{equation}
\label{distdiag}
	d_{k} = \frac{1}{2}(x_i-\mu_{k})^T \Lambda_k^{-1} (x_i-\mu_{k}),\ k \in [1, K]
\end{equation}

The gradient computations for the GM loss of the $i$-th sample are given in Eqs.~\ref{eq_dmu1} to~\ref{eq_dx}.
For conciseness, we denote $p(k|x_i)$ as $p_k$ and $(x_i-\mu_{k})(x_i-\mu_{k})^T$ as $C_k$ in all these equations.
\begin{equation}
\label{eq_dmu1}
\frac{\partial \mathcal{L}_{GM,i}^m}{\partial \mu_{z_i}} = \big[\big(1 - p_{z_i}\big)(1+\alpha) + \lambda\big] \Lambda_{z_i}^{-1} (\mu_{z_i} - x_i)
\end{equation}
\begin{equation}
\label{eq_dmu2}
\frac{\partial \mathcal{L}_{GM,i}^m}{\partial \mu_k} = p_k \Lambda_{k}^{-1} (x_i - \mu_k), \ \ \forall k \neq z_i
\end{equation}
\begin{eqnarray}
\begin{aligned}
\label{eq_dsigma1}
\frac{\partial \mathcal{L}_{GM,i}^m}{\partial \Lambda_{z_i}} =& -\frac{1}{2}\big[\big((1-p_{z_i})(1+\alpha)+\lambda \big)\Lambda_{z_i}^{-1}C_k - \\
& (1-p_{z_i}+\lambda)I\big]\Lambda_{z_i}^{-1}
\end{aligned}
\end{eqnarray}
\begin{equation}
\label{eq_dsigma2}
\frac{\partial \mathcal{L}_{GM,i}^m}{\partial \Lambda_{k}} = -\frac{1}{2}p_k (I- \Lambda_k^{-1}C_k)\Lambda_k^{-1}, \ \ \forall k \neq z_i
\end{equation}
\begin{eqnarray}
\label{eq_dx}
\begin{aligned}
\frac{\partial \mathcal{L}_{GM,i}^m}{\partial x_i} = &\big[\big(1 - p_{z_i}\big)(1+\alpha) + \lambda\big]\Lambda_{z_i}^{-1} (x - \mu_{z_i}) \\
&- \sum_{k \ne z_i} p_k\Lambda_k^{-1} (x_i - \mu_k)
\end{aligned}
\end{eqnarray}

\section{Experiments}
\label{exp}

Two sets of experiments are presented in this section.
In the first set, we conduct the image classification and face verification experiments to verify the effectiveness of the large-margin GM loss (L-GM loss for short).
We report mean and standard deviation of 3 tries.
In the second set, we demonstrate the feasibility of distinguishing adversarial examples using the likelihood regularization term $\mathcal{L}_{lkd}$.
All experiments are carried out using the \emph{Caffe} framework~\cite{jia2014caffe} on NVIDIA TitanX GPUs. 

For the margin  parameter $\alpha$, a larger value may lead to a more difficult optimization objective.
Therefore intuitively, $\alpha$ should be smaller when the number of classes gets larger.
In our experiments, we empirically set $\alpha$ to 1.0, 0.3, 0.1, 0.01 and 0.01 for MNIST, CIFAR-10, CIFAR-100, ImageNet and face verification, respectively.
Also, we set the likelihood regularization parameter $\lambda$ to a small value, \eg 0.1 in our experiments, so that the likelihood regularization starts to play a major role when the training accuracy is approaching saturation, or when $p_{z_i}$ approaches 1.

\subsection{Image Classification}
\label{sect_exp_visual}

\paragraph{MNIST} We first compare the softmax loss, the center loss (with the softmax loss) \cite{{DBLP:conf/eccv/WenZL016}}, the large-margin softmax loss (L-Softmax loss for short) \cite{DBLP:conf/cvpr/SchroffKP15} and the L-GM loss by visualizing their learned 2D feature spaces for the MNIST Handwritten Digit dataset \cite{mnist}.
We adopt a network with 6 convolution layers and a fully connected layer with a two dimensional output.
The feature embeddings on the training set with different loss functions are illustrated in Fig.~\ref{fig01}. 
As we can see, different from the softmax loss and its variants, the features generated using the L-GM loss roughly follow the GM distribution, which is consistent with the assumption.
The heatmap of the learned likelihood is shown in  Fig.~\ref{fig01}(f).
Also, as is shown in Fig.~\ref{fig01} (d)-(e), with an increasing $\alpha$, larger margin sizes can be observed among different classes. 

For the quantitative evaluation, we also increase the output dimension of the fully connected layer from 2 to 100 and add a ReLU activation after it.
For fair comparison, we train the same network with different loss functions using identical training parameters including the learning rate, weight decay, etc..
The classification accuracies on the test set are presented in Table~\ref{tb_mnist}.

\begin{table}[t]
	\centering
	\begin{tabular}{c|cc}
		\hline
		Loss Functions & 2-D (\%) & 100-D (\%) \\
		\hline
		Center \cite{{DBLP:conf/eccv/WenZL016}}  & 1.45 $\pm$ 0.01 & 0.47 $\pm$ 0.01 \\
		L-Softmax\cite{DBLP:conf/icml/LiuWYY16} & 1.30 $\pm$ 0.02 & 0.43 $\pm$ 0.01 \\
		Softmax & 1.82 $\pm$ 0.01 & 0.68 $\pm$ 0.01 \\
		\hline
		L-GM ($\alpha=0$))      & 1.44 $\pm$ 0.01 & 0.49 $\pm$ 0.01\\
	    L-GM ($\alpha=0.3$)  & 1.32 $\pm$ 0.01 & 0.42 $\pm$ 0.02\\
		L-GM ($\alpha=1.0$) & \textbf{1.17 $\pm$ 0.01} & \textbf{0.39 $\pm$ 0.01} \\
		\hline
	\end{tabular}
	\caption{Recognition error rates (\%) on MNIST test set using a 6-layer CNN with different loss functions.}
\label{tb_mnist}
\end{table}

\paragraph{CIFAR} CIFAR-10 and CIFAR-100 \cite{krizhevsky2009learning} each consists of of $32\times 32$ pixel colored images, with 50,000 training images and 10,000 testing images.
We adopt the standard data augmentation scheme including mirroring and $32\times 32$ random cropping after 4 pixel zero-paddings on each side \cite{DBLP:conf/cvpr/HeZRS16, DBLP:conf/icml/LiuWYY16}.

For CIFAR-10, We train the ResNet \cite{DBLP:conf/cvpr/HeZRS16} of depth 20, 56 and 110 with different loss functions. 
The networks are trained with a batch size of 128 for 300 epochs; and the learning rate is set to 0.1 and then divided by 10 at the $150^{th}$ epoch and the $225^{th}$ epoch respectively. 
We use a weight decay of $5\times 10^{-4}$ and the Nesterov optimization algorithm \cite{sutskever2013importance} with a momentum of 0.9.
The network weights are initialized using the method introduced in \cite{he2015delving}.
The recognition accuracies are shown in Table~\ref{tb_cifar10}.
Results in the first row were reported in the original RestNet paper \cite{DBLP:conf/cvpr/HeZRS16}.
For the center loss and the large-margin softmax loss, we train the models by ourselves since the ResNet was not used on CIFAR-10 in the original papers \cite{DBLP:conf/eccv/WenZL016} and \cite{DBLP:conf/icml/LiuWYY16}. 
The proposed L-GM loss outperforms the softmax loss and its two variants for different ResNet models with various depths.

\begin{table}[tb]
\small
	\centering
	\begin{tabular}{c|c|c|c}
		\hline
		Loss Functions & ResNet-20 & ResNet-56 & ResNet-110 \\
		\hline
		\hline
		Softmax \cite{DBLP:conf/cvpr/HeZRS16} & 8.75 $\pm$ 0.04 & 6.97 $\pm$ 0.05 & 6.43 $\pm$ 0.04 \\
		Center \cite{DBLP:conf/eccv/WenZL016} & 7.77 $\pm$ 0.05 & 5.94 $\pm$ 0.02 & 5.32 $\pm$ 0.03 \\
		L-Softmax \cite{DBLP:conf/icml/LiuWYY16} & 7.73 $\pm$ 0.03 & 6.05 $\pm$ 0.04 & 5.79 $\pm$ 0.02 \\
		\hline
		L-GM($\alpha=0.3$) & \textbf{7.21 $\pm$ 0.04} & \textbf{5.61 $\pm$ 0.02} & \textbf{4.96 $\pm$ 0.03} \\
		\hline
	\end{tabular}	
	\caption{Recognition error rates (\%) on CIFAR-10 using ResNet models with different loss functions.}
   	\label{tb_cifar10}
\end{table}

For CIFAR-100, we adopt the same CNN architecture used by the large-margin softmax loss \cite{DBLP:conf/icml/LiuWYY16}, which follows the design philosophy of the VGG-net \cite{DBLP:journals/corr/SimonyanZ14a} consisting of 13 convolutional layers and 1 fully connected layer. 
Bach normalization \cite{DBLP:journals/corr/IoffeS15} is used after each convolutional layer and no dropout is used.
To achieve better recognition performances, we replace the fully connected layer in this network with Global Average Pooling \cite{DBLP:journals/corr/LinCY13}.
We report the recognition performances with or without the data augmentation in Table \ref{tb_cifar100}, denoted by C100+ and C100 respectively.
Several points can be observed from Table \ref{tb_cifar100}.
First, the proposed L-GM loss consistently outperforms the softmax based losses on both C100+ and C100.
Second, for the augmented data set C100+, increasing the margin parameter $\alpha$ consistently benefits the recognition performance. However, this is not true for C100 without data augmentation.
This is probably related to the fact that the number of training samples for each object class is as low as 500 on C100.
The margin size on the training set and the model generalization capability is less correlated.

\begin{table}[h]
	\centering
	\begin{tabular}{c|c|c}
		\hline
		Loss Functions & C100 & C100+ \\
		\hline
		\hline
		Center \cite{{DBLP:conf/eccv/WenZL016}} & 24.85 $\pm$ 0.06 & 21.05 $\pm$ 0.03\\
		L-Softmax \cite{DBLP:conf/icml/LiuWYY16} & 24.83 $\pm$ 0.05 & 20.98 $\pm$ 0.04 \\
		Softmax & 25.61 $\pm$ 0.07 & 21.60 $\pm$ 0.04 \\
		\hline
		LGM($\alpha=0.1$)  & 23.74 $\pm$ 0.08 & 20.94 $\pm$ 0.03\\
		LGM($\alpha=0.2$)  & \textbf{23.04 $\pm$ 0.08} & 20.85 $\pm$ 0.04 \\
		LGM($\alpha=0.3$)  & 23.80 $\pm$ 0.06 & \textbf{20.76 $\pm$ 0.03} \\
		\hline 
	\end{tabular}
	\caption{Recognition error rates (\%) on CIFAR-100 using a VGG-like 13 layer CNN with different loss functions.}
    \label{tb_cifar100}
\end{table}

\paragraph{ImageNet} We investigate the performance on large-scale image classification using the ImageNet dataset \cite{deng2009imagenet}.
We perform experiments on ImageNet (ILSVRC2012) using ResNet-101 \cite{DBLP:conf/cvpr/HeZRS16} combined with different loss functions. 
To make fair comparison, all the models are trained for 100 epochs on 6 Titan GPUs with a mini-batch size of 16 for each GPU.
The learning rate is initialized as 0.01 and divided by 10 at the $50^{th}$ epochs and $75^{th}$ epochs respectively.
We use a weight decay of 0.0002 and a momentum of 0.9; and no dropout \cite{dropout} is used.
We evaluate the performances for 1-crop and 10-crop practices on the ILSVRC2012 validation set. Results in Table \ref{tb_imagenet} show that our proposal is also effective on the large-scale dataset.

\begin{table}[h] 
\small
  \centering  
    \begin{tabular}{c|c|c|c|c}  
    \hline  
    \multirow{2}{*}{Loss}&  
    \multicolumn{2}{c|}{1-crop}&\multicolumn{2}{c}{10-crop}\cr\cline{2-5}  
    &top-1&top-5&top-1&top-5\cr  
    \hline  
    \hline      Softmax&23.5$\pm$0.2&7.55$\pm$0.08&22.6$\pm$0.2&6.92$\pm$0.04\cr\hline
   L-GM& \textbf{22.7$\pm$0.2}& \textbf{7.14$\pm$0.08}& \textbf{21.9$\pm$0.1}& \textbf{6.05$\pm$0.03}\cr
    \hline  
    \end{tabular} 
	\caption{Error rates (\%) on ILSVRC2012 validation set. For L-GM, we set $\alpha$=0.01 and $\lambda$=0.1.}
   	\label{tb_imagenet}
\end{table} 

\subsection{Face Verification}


We conduct the face verification experiments on the Labeled Face in the Wild (LFW) dataset \cite{huang2007labeled}, which contains 13,233 face images from $5749$ different identities with large variations in pose, expression and illumination.
The officially provided 6,000 pairs are used for face verification test.
We follow the standard \emph{unrestricted, labeled outside data} protocol of LFW and use only the CASIA-WebFace dataset \cite{DBLP:journals/corr/YiLLL14a} for training. 
The CASIA-WebFace dataset consists of 494,414 face images from 10,575 subjects.
The training and testing images are aligned using MTCNN \cite{zhang2016joint} and resized to $128\times 128$ pixel.
A simple data augmentation scheme is adopted including horizontal mirroring and $120\times 120$ random crop from the aligned $128\times 128$ pixel face images.

\begin{table}[b]
	\centering
	\begin{tabular}{c|c|c}
		\hline
		Method & Training Data & Accuracy \\
		\hline
		\hline
		FaceNet \cite{schroff2015facenet} & 200M & \textbf{99.65} \\
		Deepid2+ \cite{sun2015deeply} & 0.3M & 98.70 \\ 
		\hline
		Softmax & 0.49M & 98.56 $\pm$ 0.03\\
		L-Softmax \cite{DBLP:conf/icml/LiuWYY16} & 0.49M & 98.92 $\pm$ 0.03\\
		Center \cite{DBLP:conf/eccv/WenZL016} & 0.49M & 99.05 $\pm$ 0.02\\
		LGM ($\alpha=0.001$) & 0.49M & 99.03 $\pm$ 0.03\\
		LGM ($\alpha=0.005$) & 0.49M & 99.08 $\pm$ 0.02\\
		LGM ($\alpha=0.01$) & 0.49M & \textbf{99.20 $\pm$ 0.03} \\
		\hline
	\end{tabular}
	\caption{Face verification performances on LFW of a single model. The 6 models at bottom are trained on our scheme while the 2 results on top are reported from the original paper.}
	\label{tb_face}
\end{table}

We train the ResNet \cite{DBLP:conf/cvpr/HeZRS16} based face recognition model  with 27 convolutional layers. 
The PReLU activations \cite{he2015delving} are used after each convolutional layer and no batch normalization or Dropout is used.
We train with a batch size of 256 for 20 epochs.
The learning rate is initially set to 0.1 and divided by 10 at the 10th, 14th and 16th epochs.
The networks are trained using stochastic gradient descent (SGD) with a momentum of 0.9 and a weight decay of $5\times 10^{-4}$.

For the L-GM loss, we perform PCA on the 512-dimensional feature embeddings and then compute the Mahalanobis distance for verification.
For fair comparison, the verification performance is evaluated on single models and model ensemble is not used. 
In Table \ref{tb_face}, the accuracies for the Deepid2+ (contrastive loss) \cite{sun2015deeply} and the FaceNet (triplet loss) \cite{schroff2015facenet} are reported in the original papers.
The FaceNet achieves the highest accuracy of 99.65\% by using a very large training set of 200M images.
In \cite{DBLP:conf/eccv/WenZL016}, Y. Wen \etal reported a higher accuracy of 99.28\% for the center loss by using both the CASIA-Webface and the Celebrity+ \cite{Liu2014Deep} dataset for training, with 0.7M training images in total.
When using the CASIA-Webface training dataset only, the L-GM loss outperforms the other loss functions.


%
%

\subsection{Beyond Classification}

As we have discussed in Sect.~\ref{discussion}, the proposed L-GM loss enables the likelihood estimation for a given input in addition to the class prediction.
During training, the L-GM loss drives the deep model to generate features that follow the assumed GM distribution as well as possible, while guaranteeing the inter class separability.
In other words, the training feature distribution is supposed to be well established for a trained deep model using the L-GM loss.
We will validate this claim through experiments on distinguishing adversarial examples from normal inputs in this section.

\paragraph{Adversarial Examples} 

For a deep neural network, adversarial examples are inputs formed by intentionally adding small but worst-case perturbations which cause the model to make incorrect classifications with high confidence ~\cite{goodfellow2014explaining}.
We generate the adversarial examples using the fast gradient sigh method (FGSM)~\cite{goodfellow2014explaining}, which uses gradient backpropagation to perturb the inputs so as to maximize the classification loss.
The perturbation $P$ is generated by $P = \epsilon \cdot sign(\nabla_I \mathcal{L}(I,z))$, in which $\mathcal{L}$ is the classification loss function (\eg $\mathcal{L}_{cls}$ in L-GM loss), $I$ is the input image, $z$ is the true class label, and $\epsilon>0$ is called the magnitude of perturbation.
Then the adversarial example is formed by adding $P$ to the original image $I$.
An extension of FGSM called the Targeted FGSM aims at misclassifying an input sample to a target class by minimizing the loss for the pre-set target label $\tilde{z}$.
The targeted perturbation $P_t$ is given by
$P_t = \epsilon \cdot sign(-\nabla_I \mathcal{L}(I,\tilde{z}))$, in which $\mathcal{L}$ is $\mathcal{L}_{GM}$ in our experiments.

\begin{figure}[t]
	\centering
	\includegraphics[width=0.99\linewidth]{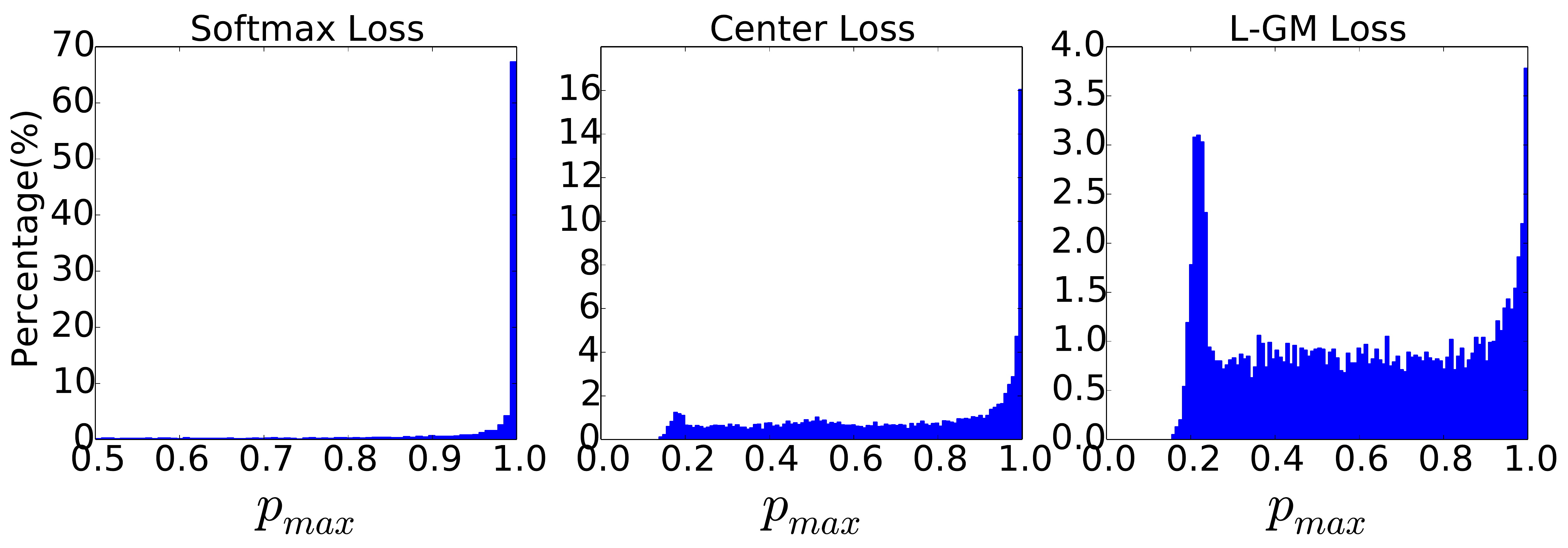}
	\caption{Histograms of the predicted posterior probability of the adversarial examples.}
   \label{fig_prob}
\end{figure}

By using the FGSM, we first generate one adversarial example for each MNIST test image in order to evaluate the classification performance using different loss functions.
As such there are altogether 10,000 adversarial examples and 10,000 original normal MNIST test images in the experiment.
We use the CNN architecture as described in Sect.~\ref{sect_exp_visual} with the 100-dimensional feature embedding.
Three models are trained on the standard MNIST training set by using the softmax loss, the center loss and the proposed L-GM loss respectively.
The classification error rates on the adversarial examples are presented in Table~\ref{tb_mnist_adv}, which shows that all three models seem to be vulnerable to adversarial attacks.
We then investigate the posterior probability ($p_{max} = \max_k p(k|x)$) corresponding to the predicted class for both the normal inputs and the adversarial examples ($\epsilon=0.3$).
For adversarial examples, the histograms of $p_{max}$ are shown in Fig.~\ref{fig_prob}.
For normal inputs, it is unnecessary to plot the histograms since the $p_{max}>0.98$ for over 95\% of the samples for all the three losses.
Obviously, for the L-GM loss, the overlap between the histograms of $p_{max}$ of the normal inputs and the adversarial examples is the smallest among three loss functions.
This means that even by only considering the posterior probability in classification, the L-GM loss already outperforms the other two loss functions in distinguishing adversarial examples.
Nevertheless, a more effective way for distinguishing adversarial examples is to directly consider the likelihood to the learned training feature distribution.
We therefore design the following experiment. 

\begin{table}[t]
	\centering
	\begin{tabular}{c|c|c|c}
		\hline
		$\epsilon$ & Softmax  & Center & L-GM($\alpha=1$) \\
		\hline
        \hline
		0 & 0.68 & 0.47 & 0.39 \\
		0.1 & 24.08 & 43.13  & 23.63 \\
		0.2 &  75.56 &  67.17  &  64.40 \\
        0.3 & 84.87 & 85.49 &  81.62 \\
		\hline
	\end{tabular}
	\caption{Classification error rates (\%) on adversarial examples generated from the MNIST test set using FGSM. $\epsilon=0$ means that the inputs are normal MNIST test images.}
\label{tb_mnist_adv}
\end{table}

\paragraph{Adversarial Verification}

Intuitively, in the feature space, the adversarial examples should follow a distribution different from that of the normal inputs.
Based on this understanding, we design an experiment called the \textit{adversarial verification} to distinguish the adversarial examples from normal inputs based on the feature likelihood.
Let the predicted class be $\hat{z}_i = \mathop{\arg\max}_{k} \ p(k|x_i)$.
For the L-GM loss, we now assume identity covariance matrix and equal priors for simplicity. 
Then the likelihood of $x_i$ is $l_{GM,i}=exp(-\|x_i-\mu_{\hat{z}_i}\|^2/2)$ based on Eq.~\ref{likelihood} by omitting the constant coefficient.
And it can also be rewritten as $l_{GM,i}=exp(-\mathcal{L}_{lkd,i})$ according to Eq.\ref{l_reg}.
For the center loss, the likelihood can be computed similarly according to Lemma ~\ref{theo1}, leading to $l_{C,i} = exp(-\|x_i-\mu_{\hat{z}_i}\|^2/2)$.
For the softmax loss, the likelihood is not explicitly established in its formulation.
A reasonable way is to estimate the likelihood as $l_{S,i}=w_{\hat{z}_i}^Tx_i + b_{\hat{z}_i}$. After all, the affinity score $w_{\hat{z}_i}^Tx_i + b_{\hat{z}_i}$ represents the similarity between $x_i$ and class $\hat{z}_i$.

\begin{figure}[t]
	\centering
	\includegraphics[width=0.99\linewidth]{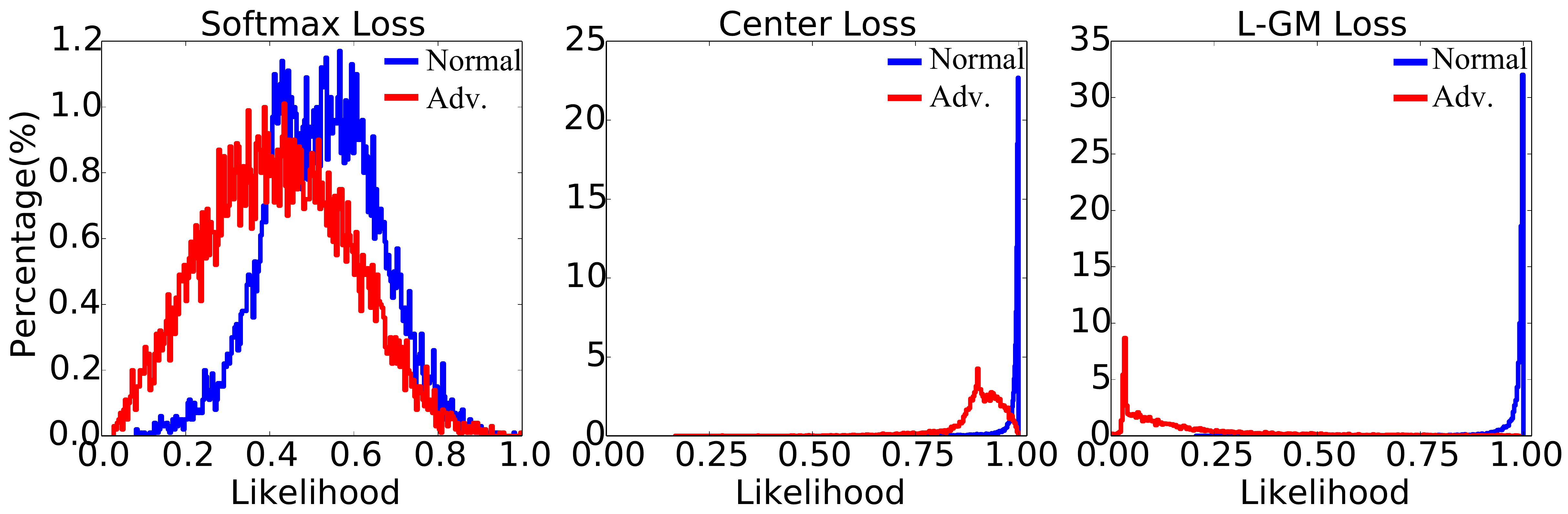}
	\caption{Histograms of the likelihood for adversarial examples (Adv.) and normal inputs (Normal). }
   \label{fig_likelihood}
\end{figure}

In the adversarial verification experiment, the FGSM is used to generate adversarial examples for the MNIST test set, with $\epsilon=0.3$. 
Then for the three models, we compute the likelihood of the normal test images and the adversarial examples.
For the softmax loss, we normalize the likelihood $l_S$ to (0, 1] for comparison.
The histograms of the likelihood for three loss functions are illustrated in Fig.~\ref{fig_likelihood}.
For the L-GM loss, the adversarial examples have very low likelihood in the feature space and the normal inputs can be easily distinguished from them.
The softmax loss, however, clearly suffers from a serious overlap between the two likelihood histograms.
The center loss lies in between by being superior to the softmax loss while inferior to the L-GM loss in terms of the capability of adversarial verification.

\begin{figure}[t]
	\centering
	\includegraphics[width=0.7\linewidth]{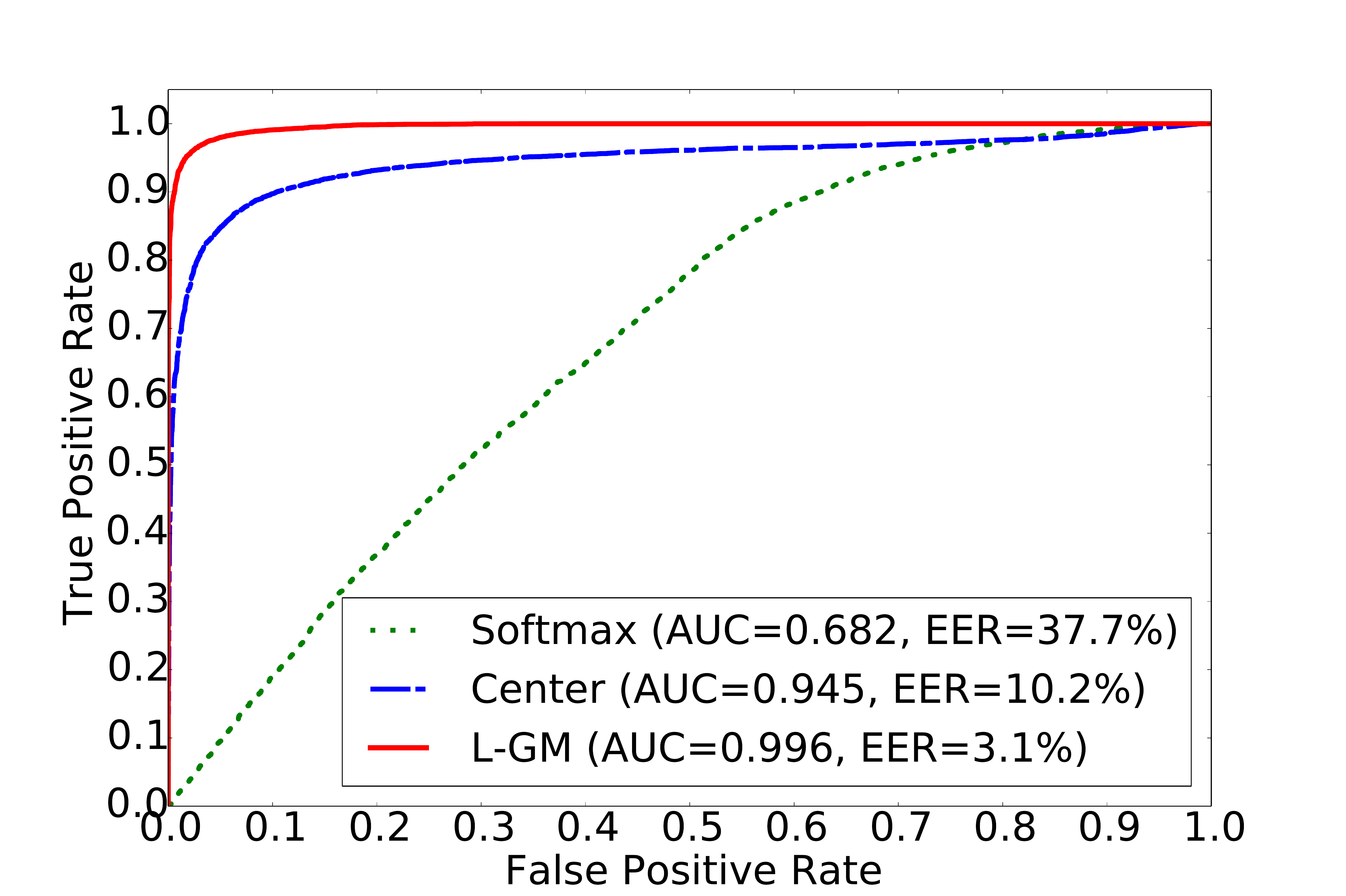}
	\caption{ROC curves of the adversarial verification.}
   \label{fig_roc}
\end{figure}

Quantitatively, we evaluate the adversarial verification performances by thresholding the likelihood, and resultant ROC curves are demonstrated in Fig.~\ref{fig_roc}.
The equal error rate (EER) for the softmax loss is 37.7\%, which is practically too high in a binary classification task.
The center loss performs much better with an EER of 10.2\%.
The proposed L-GM loss achieves the lowest EER of 3.1\%.
This experiment demonstrates that comparing to the other two loss functions, the L-GM loss can be effectively used for distinguishing adversarial examples.
This validates our claim that the L-GM loss can well establish the training feature distribution while maintaining a satisfactory classification performance.

\paragraph{Discussions}

\begin{figure}[t]
	\centering
	\includegraphics[width=0.55\linewidth]{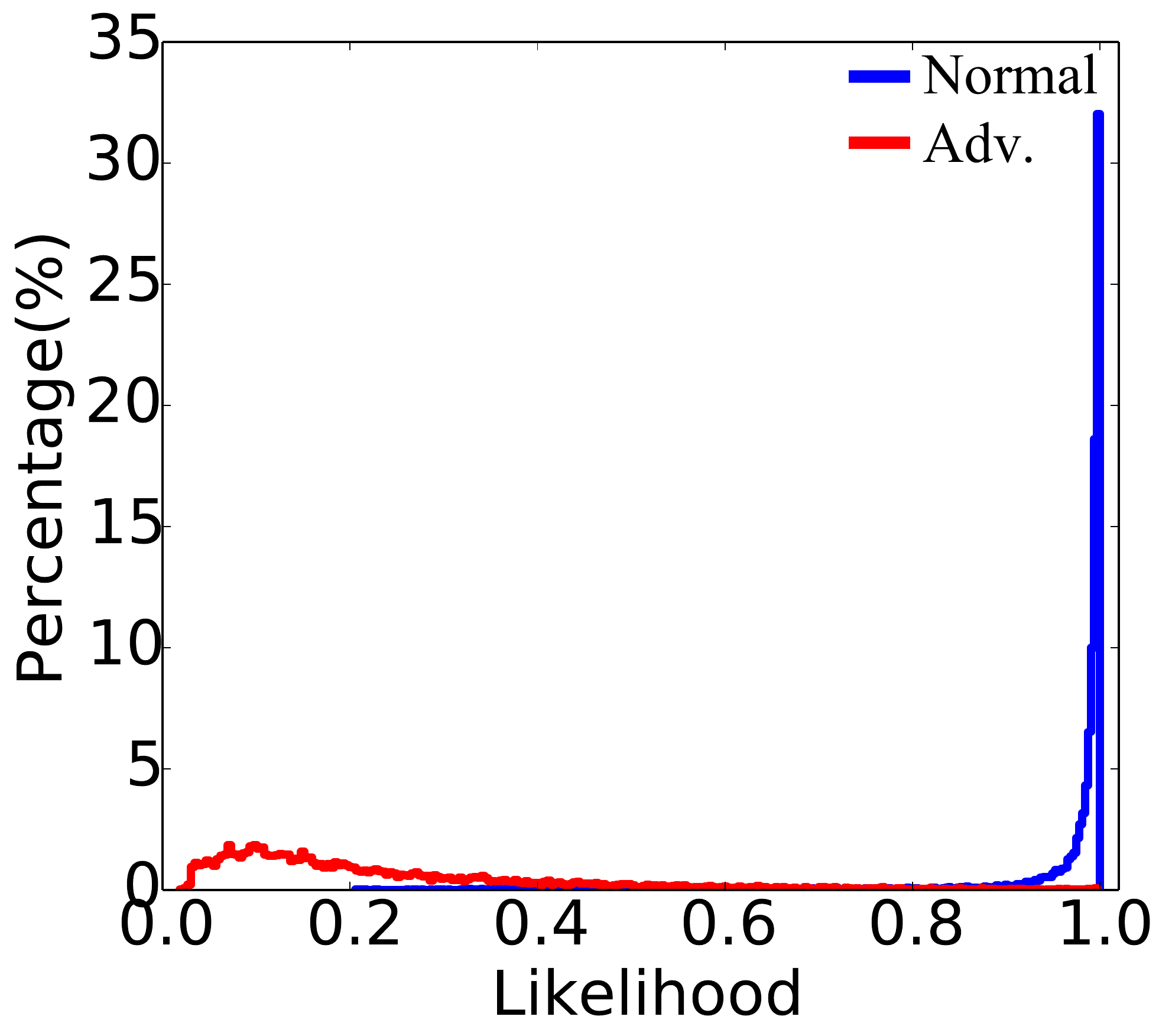}
	\caption{Histogram of the likelihood for adversarial examples generated by the Targeted FGSM against the L-GM loss.}
   \label{fig_likelihood_target}
\end{figure}

Theoretically speaking, it is possible to generate adversarial examples with high likelihood in the L-GM loss by jointly optimizing the classification loss and the likelihood regularization term.
It can be verified that under the L-GM loss formulation, such a joint optimization can be approximately realized using the Targeted FGSM, in which the targeted perturbation $P_t$ actually helps to reduce the distance between the feature and the center of the targeted class, or increase the likelihood.
We test this approach by using the class with the second largest posterior probability as the target label $\tilde{z}$ for a given input.
We still set $\epsilon=0.3$ and only test the L-GM loss.
The classification error rate is 81.37\%, which is similar to that in Table.~\ref{tb_mnist_adv}.
The likelihood histogram is illustrated in Fig.~\ref{fig_likelihood_target}.
Compared to Fig.~\ref{fig_likelihood}, the number of adversarial examples with very low likelihood (\eg smaller than 0.2) is decreased, leading to a slightly higher EER of 4.3\%.
Nevertheless, most of the adversarial examples can still be distinguished using the likelihood.

\section{Conclusions}
\label{conclusions}

We proposed a loss function by assuming a Gaussian Mixture (GM) distribution of the deep features on the training set.
Besides the classification loss, a log likelihood regularization term is added to explicitly drive the deep model for generating GM distributed features.
To further improve the generalization capability of the trained model, a classification margin is introduced. 
Extensive experiments demonstrate that the proposed L-GM loss outperforms the softmax loss and its variants in in both small and large-scale datasets when combined with different deep models.
Besides, the L-GM loss facilitates a more effective distinguishment of abnormal inputs of which the extracted features follow a distribution different from the one learned during training.
This can be practically useful, for example, to improve an deep model's robustness towards adversarial examples.
\paragraph{Acknowledgements} 
This work was supported by the National Natural Science Foundation of China (61673234).

{\small
\bibliographystyle{ieee}
\bibliography{reference}
}

\end{document}